# CrossLight: A Cross-Layer Optimized Silicon Photonic Neural Network Accelerator

Febin Sunny, Asif Mirza, Mahdi Nikdast, and Sudeep Pasricha
Colorado State University, Fort Collins, CO, USA
{febin.sunny, mirza.baig, mahdi.nikdast, sudeep}@colostate.edu

*Abstract* – **Domain-specific neural network accelerators have seen growing interest in recent years due to their improved energy efficiency and inference performance compared to CPUs and GPUs. In this paper, we propose a novel cross-layer optimized neural network accelerator called *CrossLight* that leverages silicon photonics. *CrossLight* includes device-level engineering for resilience to process variations and thermal crosstalk, circuit-level tuning enhancements for inference latency reduction, and architecture-level optimization to enable higher resolution, better energy-efficiency, and improved throughput. On average, *CrossLight* offers 9.5× lower energy-per-bit and 15.9× higher performance-per-watt at 16-bit resolution than state-of-the-art photonic deep learning accelerators.**

## I. INTRODUCTION

Many emerging applications such as self-driving cars, autonomous robotics, fake news detection, pandemic growth and trend prediction, and real-time language translation are increasingly being powered by sophisticated machine learning models. With researchers creating deeper and more complex deep neural network (DNN) architectures, including multi-layer perceptron (MLP) and convolution neural network (CNN) architectures, the underlying hardware platform must consistently deliver better performance while satisfying strict power dissipation limits. This endeavor to achieve higher performance-per-watt has driven hardware architects to design custom accelerators for deep learning, e.g., Google's TPU [1] and Intel's Movidius [2], with much higher performance-per-watt than conventional CPUs and GPUs.

Unfortunately, electronic accelerator architectures face fundamental limits in the post Moore's law era where processing capabilities are no longer improving as they did over the past several decades [3]. In particular, moving data electronically on metallic wires in these accelerators creates a major bandwidth and energy bottleneck [4]. Silicon photonics is a promising technology to enable ultra-high bandwidth, low-latency, and energy-efficient communication solutions [5]. CMOS-compatible photonic interconnects have already replaced metallic ones for light-speed data transmission at almost every level of computing, and are now actively being considered for chip-scale integration [6].

Remarkably, it is also possible to use optical components to perform computation, e.g., matrix-vector multiplication [7]. Thus, it is now possible to conceive of a new class of DNN accelerators that employ photonic interconnects and photonic integrated circuits (PICs) built with on-chip waveguides, electro-optic modulators, photodetectors, and lasers for low-latency and energy-efficient optical domain data transport and computation. Not only can such photonics-based accelerators address the fan-in and fan-out problems with linear algebra processors, but their operational bandwidth can approach the photodetection rate (typically in the hundreds of GHz), which is orders of magnitude higher than electronic systems today that operate at a clock rate of a few GHz [8].

Despite the above benefits, a number of obstacles must be overcome before viable photonic DNN accelerators can be realized. Fabrication process and thermal variations can adversely impact the robustness of photonic accelerator designs by introducing undesirable crosstalk noise, optical phase shifts, resonance drifts, tuning overheads, and photo-detection current mismatches. For example, experimental studies have shown that micro-ring resonator (MR) devices used in chip-scale photonic interconnects can experience significant resonant drifts (e.g., ~9 nm reported in [9]) within a wafer due to process variations. This matters because even a 0.25 nm drift can cause the bit-error-rate (BER) of photonic data traversal to degrade from $10^{-12}$ to $10^{-6}$. Moreover, thermal crosstalk in silicon photonic devices such as MRs can limit the achievable precision (i.e., resolution) of weight and bias parameters to a few bits, which can significantly reduce DNN model accuracy. Common tuning circuits that rely on thermo-optic phase-change effects to control photonic devices, e.g., when imprinting activations or weights on optical signals, also place a limit on the achievable throughput and parallelism in photonic accelerators. Lastly, at the architecture level, there is a need for a scalable, adaptive, and low-cost computation and communication fabric that can handle the demands of diverse MLP and CNN models.

In this paper, we introduce *CrossLight*, novel silicon photonic neural network accelerator that addresses the challenges highlighted above through a cross-layer design approach. By cross-layer, we refer to the design paradigm that involves considering multiple layers in the hardware-software design stack together, for a more holistic optimization of the photonic accelerator. *CrossLight* involves device-level engineering for resilience to fabrication-process variations and thermal crosstalk, circuit-level tuning enhancements for inference latency reduction, and an optimized architecture-level design that also integrates the device- and circuit-level improvements to enable higher resolution, better energy-efficiency, and improved throughput compared to prior efforts on photonic accelerator design. Our novel contributions in this work include:

- Improved silicon photonic device designs that we fabricated to make our architecture more resilient to fabrication-process variations;
- An enhanced tuning circuit to simultaneously support large thermal-induced resonance shifts and high-speed, low-loss device tuning;
- Consideration of thermal crosstalk mitigation methods to improve the weight resolution achievable by our architecture;
- Improved wavelength reuse and use of matrix decomposition at the architecture-level to increase throughput and energy-efficiency;
- A comprehensive comparison with state-of-the-art accelerators that shows the efficacy of our cross-layer optimized solution.

## II. BACKGROUND AND RELATED WORK

Silicon-photonics based DNN accelerator architectures represent an emerging paradigm that can immensely benefit the landscape of deep learning hardware design [10]-[14]. A photonic neuron in these architectures is analogous to an artificial neuron and consists of three components: a weighting, a summing, and a nonlinear unit. Noncoherent photonic accelerators, such as [11]-[13], typically employ the Broadcast and Weight (B&W) protocol [10] to manipulate optical signal power for setting and updating weights and activations. The B&W protocol is an analog networking protocol that uses wavelength-division multiplexing (WDM), photonic multiplexors, and photodetectors to combine outputs from photonic neurons in a layer. Coherent photonic accelerators, such as [8], [14], manipulate the electrical field amplitude rather than signal power and typically use only a single wavelength. Weighting occurs with electrical field amplitude attenuation proportional to the weight value, and phase modulation that is proportional to the sign of the weight. The weighted signals are then coherently accumulated with cascaded Y-junction combiners. For both types of accelerators, non-linearity can be implemented with devices such as electro-absorption modulators [8].

Due to the scalability, phase encoding noise, and phase error accumulation limitations of coherent accelerators [15], there is growing interest in designing efficient noncoherent photonic accelerators. In particular, the authors of DEAP-CNN [11] have described a noncoherent neural network accelerator that implements the entirety of the CNN layers using connected convolution units. In these units, the tuned MRs assume the kernel values by using phase tuning to manipulate the energy in their resonant wavelengths. Holylight [12] is another noncoherent architecture that uses microdisks (instead of MRs) for its lower area and power consumption. It utilizes a "whispering gallery mode" resonance for microdisk operation, which unfortunately is inherently lossy due to a phenomenon called tunneling ray attenuation [16]. More generally, these noncoherent architectures suffer from susceptibility to process variations and thermal crosstalk, which are not addressed in these architectures. Microsecond-granularity thermo-optic tuning latencies further reduce the speed and efficiency of optical computing [17]. We address these shortcomings as part of our proposed cross-layer optimized noncoherent photonic accelerator architecture in this work.

### III. NONCOHERENT PHOTONIC COMPUTATION OVERVIEW

As mentioned earlier, noncoherent photonic accelerators typically utilize the Broadcast and Weight (B&W) photonic neuron configuration with multiple wavelengths. Fig. 1 shows an example of this B&W configuration with $n$ neurons in a layer where the colored-dotted box represents a single neuron. Each input to a neuron is imprinted onto a unique wavelength ($\lambda_i$) emitted by a laser diode (LD) using a Mach–Zehnder modulator (MZM). The wavelengths are multiplexed (MUXed) into a single waveguide using arrayed waveguide grating (AWG), and split into $n$ branches that are each weighted with a micro-ring resonator (MR) bank that alters optical signal power proportional to weight values. A balanced photodetector (BPD) performs summation across positive and negative weight arms at each branch. Optoelectronic devices such as electro-absorption modulators (not shown for brevity) introduce non-linearity after the multiplication and summation operations.

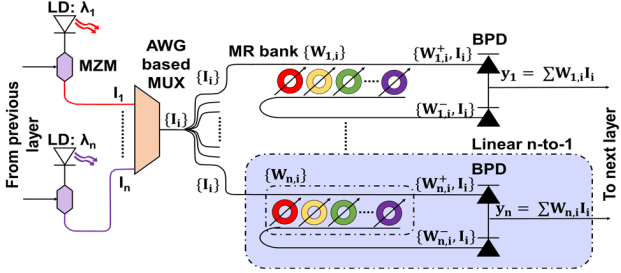

Fig. 1: Noncoherent Broadcast-and-weight (B&W) based photonic neuron.

MRs are the fundamental components that impact the efficiency of this configuration. Weights (and biases) are altered by tuning MRs so that the losses experienced by wavelengths—on which activations have been imprinted—can be modified to realize matrix-vector multiplication. MR-weight banks have groups of these tunable MRs, each of which can be tuned to drain energy from a specific resonant wavelength so that the intensity of the wavelength reflects a specific value (after it has passed near the MR). As an example of performing computation in the optical domain, consider the case where an activation value of 0.8 must be weighted by a value of 0.5 as part of a matrix-vector multiplication in a DNN model inference phase. Let us assume that the red wavelength ($\lambda_1$) is imprinted with the activation value of 0.8 by using the MZM in Fig. 1 (alternatively, MRs can be used for the same goal, where an MR will be tuned in such a way that 20% of the input optical signal intensity is dropped as the wave traverses the MR). When $\lambda_1$ passes through an MR bank, e.g., the one in the dotted-blue box in Fig. 1, the MR in resonance with $\lambda_1$ can be tuned to drop 50% of the input signal intensity. Thus, as $\lambda_1$ passes this MR, we will obtain 50% of the input intensity at the through port, which is 0.4 (=0.8×0.5). The BPD shown in Fig. 1 then converts the optical signal intensity from that wavelength (and other wavelengths) into an electrical signal that represents an accumulated single value.

An MR is essentially an on-chip resonator which is said to be in resonance when an optical wavelength on the input port matches with the resonant wavelength of the MR, generating a Lorentzian-shaped signal at the through port. Fig. 2 shows an example of an all-pass MR and its output optical spectrum. The extinction ratio (ER) and free-spectral range (FSR) are two primary characteristics of an MR. These depend on several physical properties in the MR, including its width, thickness, radius, and the gap between the input and ring waveguide [18]. Changing any of these properties changes the effective index ($n_{eff}$) of the MR, which in turn causes a change in the output optical spectrum. For reliable operation of MRs, it is crucial to maintain the central wavelength at the output optical spectrum. However, MRs are sensitive to fabrication-process variations (FPVs) and variations in surrounding temperature. These cause the central wavelength of the MR to deviate from its original position, causing a drift in the MR resonant wavelength ($\Delta\lambda_{MR}$) [19]. Such a drift (due to FPV or thermal variations) can be compensated using thermo-optic (TO) or electro-optic (EO) tuning mechanisms. Both of these have their own advantages and disadvantages. EO tuning is faster (~ns range) and consumes lower power (~4 μW/nm) but with a smaller tuning range [20]. In contrast, TO tuning has a larger tunability range, but consumes higher power (~27 mW/FSR) and has higher (~μs range) latency [17].

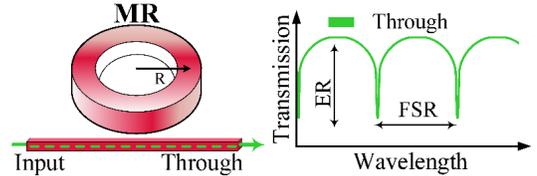

Fig. 2: An all-pass MR with output spectral characteristics at the through port with extinction ratio (ER) and free spectral range (FSR) specified in the figure.

A large number of MRs must be used at the architecture-level to support complex MLP and CNN model executions. As the number of MRs increase, so does the length of the waveguide which hosts the banks. Unfortunately, this leads to an increase in the total optical signal propagation, modulation, and through losses experienced, which in turn increases the laser power required to drive the optical signals through the weight banks, so that they can be detected error-free at the photodetector. An excessive number of parallel arms with MR weight banks (the dotted box in Fig. 1 represents one arm working in parallel with other arms) also increases optical splitter losses. Moreover, without considering crosstalk mitigation strategies (as is the case with previously proposed photonic accelerators), there is increased crosstalk noise in the optical signals, which drives down the weight resolution of the architecture.

*In summary, to design efficient photonic accelerators, there is a need for (i) improved MR device design to better tolerate variations and crosstalk; (ii) efficient MR tuning circuits to quickly and reliably imprint activation and parameter values; and (iii) a scalable architecture design that minimizes optical signal losses. Our novel Crosslight photonic accelerator design addresses all of these concerns and is discussed next.*

### IV. CROSSLIGHT ARCHITECTURE

Fig. 3 shows a high-level overview of our *CrossLight* noncoherent silicon photonic neural network accelerator. The photonic substrate performs vector dot product (VDP) operations using silicon photonic MR devices, and summation using optoelectronic photodetector (PD) devices over multiple wavelengths. An electronic control unit is required for the control of photonic devices, and for communication with a global memory to obtain the parameter values, mapping of the vectors, and for partial sum buffering. We use digital to analog converter (DAC) arrays to convert buffered signals into analog tuning signals for MRs. Analog to digital converter (ADC) arrays are used to map the output analog signals generated by PDs to digital values that are sent back for post-processing and buffering. We break down the discussion of this accelerator into three parts (subsections A-C), corresponding to the contributions at the device, tuning circuit, and architecture levels, as discussed next.

#### A. MR device engineering and fabrication

Process variations are inevitable in CMOS-compatible silicon photonic fabrications, causing undesirable changes in resonant wavelength of MR devices ($\Delta\lambda_{MR}$). We fabricated a 1.5×0.6 mm² chip with high-resolution Electron Beam (EBeam) lithography and performed a comprehensive design-space exploration of MRs to compensate for FPVs while improving MR device insertion loss and Q-factor. In this exploration, we varied the input and ring waveguide widths to find an MR device design that was tolerant to FPVs. We found that in an MR design of any radii and gap, when the input waveguide is 400 nm wide and the ring waveguide is 800 nm wide at room temperature (300 K), the undesired $\Delta\lambda_{MR}$ due to FPVs can be reduced from 7.1 to 2.1 nm (70% reduction). *This is a significant result, as these engineered MRs require less compensation for FPV-induced resonant wavelength shifts, which can reduce the power consumption of architectures using such MRs.*

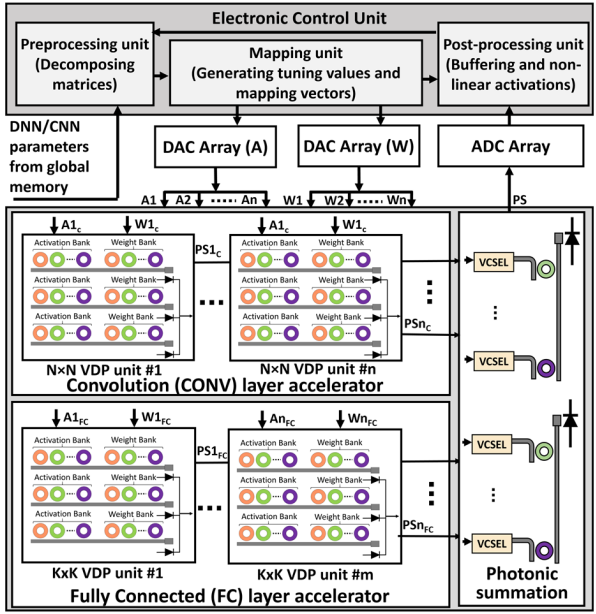

Fig. 3: An overview of *CrossLight*, showing dedicated vector dot product (VDP) units for CONV and FC layer acceleration, and the internal architecture.

Unfortunately, even with such optimized MR designs, the impact of FPVs is not completely eliminated, and there is still a need to compensate for FPVs. Thermal variations are another major factor to cause changes in MR $n_{eff}$ which also leads to undesirable $\Delta\lambda_{MR}$. Thermo-optic (TO) tuners are used to compensate for such deviations in $\Delta\lambda_{MR}$. These TO tuners use microheaters to change the temperature in the proximity of an MR device, which then alters the $n_{eff}$ of the MR, changing the device resonant wavelength, and correcting the $\Delta\lambda_{MR}$. Unfortunately, high temperatures from such heaters can cause thermal energy dissipation, creating thermal crosstalk across MR devices placed close to each other. One can avoid such thermal crosstalk by placing devices at an appropriate distance from each other, typically 120 μm to 200 μm (depending on the number of MR devices in proximity within an MR bank). But such a large spacing hurts area efficiency and also increases waveguide length, which increases propagation losses and its associated laser power overhead. We propose to address this challenge at the circuit level, as discussed next.

### B. Tuning circuit design

To reduce thermal crosstalk, we must reduce the reliance on TO tuning, an approach that is used in all prior photonic neural network accelerators, but one that entails high overheads. We propose to use a hybrid tuning circuit where both thermo-optic (TO) and electro-optic (EO) tuning are used to compensate for $\Delta\lambda_{MR}$. Such a tuning approach has previously been proposed in [22] for silicon photonic Mach–Zehnder Interferometers with low insertion loss. Such an approach can be easily transferred to an optimized MR for hybrid tuning in our architecture. The hybrid tuning approach supports faster operation of MRs with fast EO tuning to compensate for small $\Delta\lambda_{MR}$ shifts and, when necessary, using TO tuning when large $\Delta\lambda_{MR}$ shifts need to be compensated.

To further reduce the power overhead of TO tuning in this hybrid approach, we adapt a method called Thermal Eigen Decomposition (TED), which was first proposed in [23]. Using TED, we can collectively tune all the MRs in an MR bank to compensate for large $\Delta\lambda_{MR}$ shifts. By doing so, we can cancel the effect of thermal crosstalk (i.e., an undesired phase change) in MRs with much lower power consumption. The TO tuning power can be calculated by the amount of phase shift necessary to apply to the MRs in order for them to be at their desired resonant wavelength. The extent of phase crosstalk ratio (due to thermal crosstalk) as a function of the distance between an MR pair is shown in Fig. 4, for our fabricated MR devices. The results are based on detailed analysis with a commercial 3D heat transport simulation EDA tool for silicon photonic devices (Lumerical HEAT [21]). It can be seen from the orange line that as the distance between an MR pair increases, the amount of phase crosstalk reduces exponentially. Such a trend has also been observed in [24]. To find a balance between tuning power savings while having reduced crosstalk, we perform a sensitivity analysis based on the distance between two adjacent MRs in our architecture. We placed the optimized MRs (described in the previous section) in such a manner that maximum tuning power is saved when they are close to each other while compensating for thermal crosstalk. Results from our analysis (the solid-blue line in Fig. 4) indicate that placing each MR pair at a distance of 5 μm is optimal, as increasing or decreasing such a distance causes an increase in power consumption of individual TO heaters in the MRs. Fig. 4 also shows the tuning power required without using the TED approach (blue dotted line), which can be seen to be notably higher.

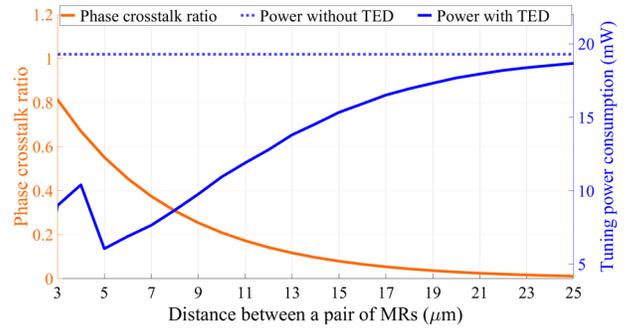

Fig. 4: Phase crosstalk ratio and tuning power consumption in a block of 10 fabricated MRs with variable distance between adjacent pair of MRs.

The workflow of our circuit-level hybrid tuning approach can be summarized as follows. When the accelerator is first booted at runtime, a one-time compensation for design-time FPVs is applied using TO tuning. The extent of compensation for crosstalk is calculated offline during the test phase, where the required phase shift in each of the MRs is calculated, and once the system is online, the respective phase shift values are applied to cancel the impact of thermal crosstalk. Subsequently, we apply EO tuning due to its extremely low latency to represent vector elements in each vector operation with MRs (discussed in more detail in the next section). If large shifts in temperature are observed at runtime, we can perform a one-time calibration with TO tuning to compensate for it. In our analysis, runtime TO tuning would be required rarely beyond its first use after the initial bootup of the photonic accelerator platform.

### C. Architecture design

The optimized MR devices, layouts, and tuning circuits are utilized within optical vector dot product (VDP) units, which are shown in Fig. 3. We use banks (groups) of MRs to imprint both activations and weights onto the optical signal. At the architecture level, we compose multiples of VDP units into two architectural sub-components: one to support convolution (CONV) layer acceleration and the other to support fully connected (FC) layer acceleration. We focus on these two types of layers as they are the most widely used and consume the most significant amount of latency and power in computational platforms that execute DNNs. In contrast, other layer types (e.g., pooling, batch normalization) can be implemented very efficiently in the electronic domain. Note also that we focus on inference acceleration, as done in all photonic DNN accelerators, and almost all electronic DNN accelerators.

*C.1 Decomposing vector operations in CONV/FC layers*

To map CONV and FC layers from DNN models to our accelerator, we first need to decompose large vector sizes into smaller ones. In CONV layers, a filter performs convolution on a patch (e.g., 2×2 elements) of the activation matrix in a channel to generate an element of the output matrix. The operation can be represented as follows:

$$K \otimes A = Y \quad (1)$$

For a 2×2 filter kernel and weight matrices, (1) can be expressed as:

$$\begin{bmatrix} k_1 & k_2 \\ k_3 & k_4 \end{bmatrix} \otimes \begin{bmatrix} a_1 & a_2 \\ a_3 & a_4 \end{bmatrix} = k_1 a_1 + k_2 a_2 + k_3 a_3 + k_4 a_4 \quad (2)$$

Rewriting (2) as a vector dot product, we have:

$$[k_1 \ k_2 \ k_3 \ k_4] \cdot \begin{bmatrix} a_1 \\ a_2 \\ a_3 \\ a_4 \end{bmatrix} = k_1 a_1 + k_2 a_2 + k_3 a_3 + k_4 a_4 \quad (3)$$

Once we are able to represent the operation as a vector dot product, it is easy to see how it can be decomposed into partial sums. For example:

$$[k_1 \ k_2] \cdot \begin{bmatrix} a_1 \\ a_2 \end{bmatrix} = k_1 a_1 + k_2 a_2 = PS_1$$

$$[k_3 \ k_4] \cdot \begin{bmatrix} a_3 \\ a_4 \end{bmatrix} = k_3 a_3 + k_4 a_4 = PS_2$$

$$PS_1 + PS_2 = Y \quad (4)$$

In FC layers, typically much larger dimension vector multiplication operations are performed between input activations and weight matrices:

$$AW = \begin{bmatrix} a_1 \\ a_2 \\ \vdots \\ a_n \end{bmatrix} [w_1 \ w_2 \ ... \ w_n] \quad (5)$$

$$AW = \begin{bmatrix} a_1 \cdot w_1 & + a_1 \cdot w_2 & + \cdots & a_1 \cdot w_n \\ a_2 \cdot w_1 & + a_2 \cdot w_2 & + \cdots & a_2 \cdot w_n \\ & & \vdots & \\ a_n \cdot w_1 & + a_n \cdot w_2 & + \cdots & a_n \cdot w_n \end{bmatrix} \quad (6)$$

In (5), $a_1$ to $a_n$ represent a column vector of activations (A) and $w_1$ to $w_n$ represent a row vector of weights (W). The resulting vector is a summation of dot products of vector elements (6). Much like with CONV layers, these can be decomposed into lower dimensional dot products.

*C.2 Vector dot product (VDP) unit design*

We separated the implementation of CONV and FC layers in *CrossLight* due to the vastly different orders of vector dot product computations required to implement each layer. For instance, typical CONV layer kernel sizes vary from 2×2 to 5×5, whereas in FC layers it is not uncommon to have 100 or more neurons (requiring 100×100 or higher order multiplication). State-of-the-art photonic DNN accelerators, e.g., [11], only consider the scales involved at the CONV layer, and either only support CONV layer acceleration in the optical domain, or use the same CONV layer implementation to accelerate FC layers. This will lead to increased latencies and reduced throughput as the larger vectors involved with FC layer calculation must be divided up into much smaller chunks, in the order of the filter kernel size of the CONV layer.

For improved efficiency, we separately support the unique scale and requirements of vector dot products involved in CONV layers and FC layers. For CONV layer acceleration, we consider *n* VDP units, with each unit supporting an *N*×*N* dot product. For FC layer acceleration, we consider *m* units, with each unit supporting a *K*×*K* dot product. Here *n>m* and *K>N*, as per the requirements of each of the distinct layers. In each of the VDP units, the original vector dimensions are decomposed into *N* or *K* dimensional vectors, as discussed above. We performed an exploration to determine the optimal values for *N, K, n,* and *m*. The results of this exploration study are presented in Section V.

*C.3 Optical wavelength reuse in VDP units*

Prior work on photonic DNN accelerator design typically considers a separate wavelength to represent each individual element of a vector. This approach leads to an increase in the total number of lasers needed in the laser bank (as the size of the vectors increases) which in turn increases power consumption. Beyond employing the decomposition approach discussed above, we also consider wavelength reuse per VDP unit to minimize laser power. In this approach, within VDP units, the *N* or *K* dimensional vectors are further decomposed into smaller sized vectors for which dot products can be performed using MRs in parallel, in each arm of the VDP unit. The same wavelengths can then be reused across arms within a VDP to reduce the number of unique wavelengths required from the laser. PDs perform summation of the element-wise products to generate partial sums from decomposed vector dot products. The partial sums from the decomposed operations are then converted back to the photonic domain by VCSELs (bottom right of Fig. 3), multiplexed into a single waveguide, and accumulated using another PD, before being sent for buffering. Thus, our approach leads to an increase in the number of PDs compared to other accelerators but significantly reduces both the number of MRs per waveguide and the overall laser power consumption.

In each arm within a VDP unit, we used a maximum of 15 MRs per bank for a total of 30 MRs per arm, to support up to a 15×15 vector dot product. The choice of MRs per arm considers not only the thermal crosstalk and layout spacing issues (discussed earlier), and the benefits of wavelength reuse (discussed in previous para), but also the fact that optical splitter losses become non-negligible as the number of MRs per arm increases, which in turn increases laser power requirements. Thus, the selection of MRs per arm within a VDP unit was carefully adjusted to balance parallelism within/across arms, and laser power overheads.

## V. EVALUATION AND SIMULATION RESULTS

*A. Simulation setup*

To evaluate the effectiveness of our *CrossLight* accelerator, we conducted several simulation studies. These studies were complemented by our MR-device fabrication and optimization efforts on real chips, as discussed in Section IV. We considered the four DNN models shown in Table I for execution on the accelerator. Model 1 is Lenet5 [25] and models 2 and 3 are custom CNNs with both FC and CONV layers. Model 4 is a Siamese CNN utilizing one-shot learning. The datasets used to train these models are also shown in the table. We designed a custom *CrossLight* accelerator simulator in Python to estimate its performance and power/energy. We used Tensorflow 2.3 along with Qkeras [26], for analyzing DNN model accuracy across different parameter resolutions.

Table I: Models and datasets considered for evaluation

| Model no. | CONV layers | FC layers | Parameters | Datasets |
|---|---|---|---|---|
| 1 | 2 | 2 | 60,074 | Sign MNIST |
| 2 | 4 | 2 | 890,410 | CIFAR10 |
| 3 | 7 | 2 | 3,204,080 | STL10 |
| 4 | 8 | 4 | 38,951,745 | Omniglot |

Table II: Parameters considered for analyses of photonic accelerators

| Devices | Latency | Power |
|---|---|---|
| EO Tuning [20] | 20 ns | 4 µW/nm |
| TO Tuning [17] | 4 µs | 27.5 mW/FSR |
| VCSEL [32] | 10 ns | 0.66 mW |
| TIA [33] | 0.15 ns | 7.2 mW |
| Photodetector [34] | 5.8 ps | 2.8 mW |

We compared *CrossLight* with the DEAP-CNN [11] and Holylight [12] photonic DNN accelerators from prior work. Table II shows the optoelectronic parameters considered for this simulation-based analysis. We considered photonic signal losses due to various factors: signal propagation (1 dB/cm [6]), splitter loss (0.13 dB [27]), combiner loss (0.9 dB [28]), MR through loss (0.02 dB [29]), MR modulation loss (0.72 dB [30]), microdisk loss (1.22 dB [31]), EO tuning loss (6 dB/cm [20]), and TO tuning loss (1 dB/cm [17]). We also considered the 1-to-56-Gb/s ADC/DAC-based transceivers from recent work [37]. To calculate laser power consumption, we use the following laser power model:

$$P_{laser} - S_{detector} \geq P_{photo\_loss} + 10 \times \log_{10} N_\lambda \quad (7)$$

where $P_{laser}$ is laser power in dBm, $S_{detector}$ is the PD sensitivity in dBm, and $P_{photo\_loss}$ is the total photonic loss encountered by the optical signal, due to all of the factors discussed above.

## B. Results: CrossLight resolution analysis

We first present an analysis of the resolution that can be achieved with *CrossLight*. We consider how the optical signals from MRs impact each other due to their spectral proximity, also known as inter-channel crosstalk. For this, we use the equations from [35]:

$$\varphi(i,j) = \frac{\delta^2}{(\lambda_i - \lambda_j)^2 + \delta^2} \quad (8)$$

In (8), $\varphi(i,j)$ describes the noise content from the $j^{th}$ MR present in the signal from the $i^{th}$ MR. As the noise content increases, the resolution achievable with *CrossLight* will decrease. Also, $(\lambda_i - \lambda_j)$ is the difference between the resonant wavelengths of $i^{th}$ MR and $j^{th}$ MR, while $\delta\ (= \lambda_i/2Q)$ denotes the 3dB bandwidth of the MRs, with Q being the quality factor (Q-factor) of the MR being considered. The noise power component can thus be calculated as:

$$P_{noise} = \sum_{i}^{n-1} \varphi(i,j) P_{in}[i] \quad (9)$$

For unit input power intensity, resolution can then be computed as:

$$Resolution = \frac{1}{max|P_{noise}|} \quad (10)$$

From this analysis, we found that with the FSR value of 18 nm and the Q value of ~8000 in our optimized MR designs, and the wavelength reuse strategy in *CrossLight*, which allows us to have large $(\lambda_i - \lambda_j)$ values (>1 nm), our MR banks will be able to achieve a resolution of 16 bits for up to 15 MRs per bank (Section IV.C.2). This is much higher than the resolution achievable by many photonic accelerators. For instance, DEAP-CNN can only achieve a resolution of 4 bits, whereas Holylight can only achieve a 2-bit resolution per microdisk (they however combine 8 microdisks to achieve an overall 16-bit resolution). Higher resolution ensures better accuracy in inference, which can be critical in some applications. Fig. 5 shows the impact of varying the resolution across the weights and activations from 1 bit to 16 bits (we used quantization-aware training to maximize accuracy), for the four DNN models considered (Table I). It can be observed that model inference accuracy is sensitive to the resolution of weight and activation parameters. Models such as the one for STL10 are particularly sensitive to the resolution. Thus, the high resolution afforded by *CrossLight* can allow achieving higher accuracies than other photonic DNN accelerators, such as DEAP-CNN.

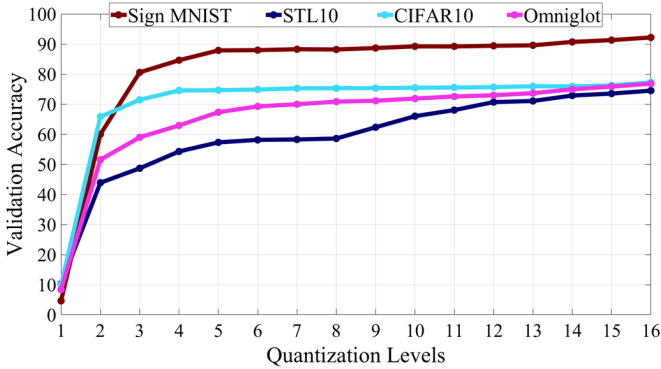

Fig. 5: Inference accuracy of the four DNN models considered, across quantization (resolution) range from 1 bit to 16 bits (for both weights and activations).

## C. Results: CrossLight sensitivity analysis

We performed a sensitivity analysis by varying the number of VDP units in the CONV layer accelerator (*n*) and FC layer accelerator (*m*), along with the complexity of the VDP units (*N* and *K*, respectively). Fig. 6 shows the frames per second (FPS; a measure of inference performance) vs. energy per bit (EPB) vs. area of various configurations of *CrossLight*. We selected the best configuration as the one that had the highest value of FPS/EPB. In terms of (*N, K, n, m*), the values of the four parameters for this configuration are (20, 150, 100, 60). This configuration also ended up being the one with the highest FPS value, but had a higher area overhead than other configurations. Nonetheless, this area is comparable to that of other photonic accelerators. We used this configuration for comparisons with prior work, as discussed next.

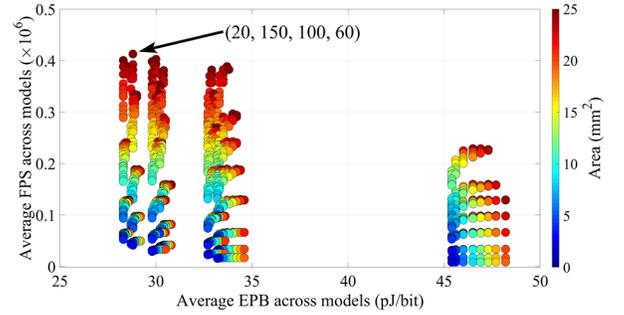

Fig. 6: Scatterplot of average FPS vs. average EPB vs. area of various *CrossLight* configurations. The configuration with highest FPS/EPB (and FPS) is highlighted.

## D. Results: Comparison with state-of-the-art accelerators

We compared our *CrossLight* accelerator against two well-known photonic accelerators: DEAP-CNN and Holylight, within a reasonable area constraint for all accelerators (~16-25 mm²). We present results for four variants of the *CrossLight* architecture: 1) *Cross_base* utilizes conventional MR designs (without FPV resilience) and traditional TO tuning; 2) *Cross_opt* utilizes the optimized MR designs from Section IV.A, and traditional TO tuning; 3) *Cross_base_TED* utilizes the conventional MR designs with the hybrid TED-based tuning approach from Section IV.B; and 4) *Cross_opt_TED* utilizes the optimized MR designs and the hybrid TED-based tuning approach.

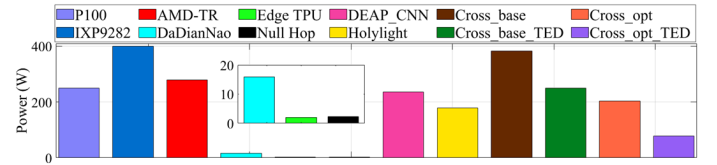

Fig. 7: Power consumption comparison among variants of *CrossLight* vs. photonic accelerators (DEAP-CNN, Holylight), and electronic accelerator platforms (P100, Xeon Platinum 9282, Threadripper 3970x, DaDianNao, EdgeTPU, Null Hop)

Fig. 7 shows the power consumption comparison across the four *CrossLight* variants and the two photonic accelerators from prior work. We also include comparison numbers for electronic platforms: three deep learning accelerators (DaDianNao, Null Hop, and EdgeTPU), a GPU (Nvidia Tesla P100), and CPUs (Intel Xeon Platinum 9282 denoted as IXP9282, and AMD Threadripper 3970x denoted as AMD-TR) [36]. The difference in power values between the *CrossLight* variants arises due to the optimization approaches adopted in each of the variant. The variants which considered conventional MR design instead of the optimized designs have larger power consumption for compensating for FPV. This value becomes non-trivial as the number of MRs increase, and thus having reduced tuning power requirement per MR (in *Cross_opt* and *Cross_opt_TED*) becomes a significant advantage. Using the TED based hybrid tuning approach provides further significant power benefits for *Cross_opt_TED* over *Cross_opt*, which uses conventional TO tuning. *Cross_opt_TED* can be seen to have lower power consumption than both photonic accelerators, as well as the CPU and GPU platforms, although this power is higher than that of the edge/mobile electronic accelerators.

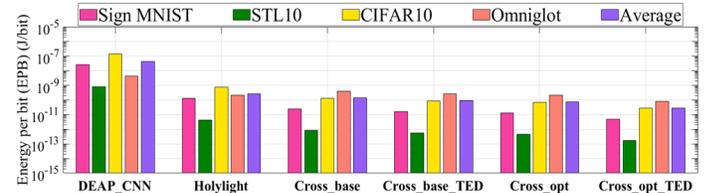

Fig. 8: Comparison of EPB values of the photonic DNN accelerators

Fig. 8 shows a comparison of energy-per-bit (EPB) across all of the photonic accelerators, for the four DNN models. On average, our best *CrossLight* configuration (*Cross_opt_TED*) has 1544× and 9.5× lower

EPB compared to DEAP-CNN and Holylight, respectively. The reason for *CrossLight*'s lower EPB is because we comprehensively took into consideration various losses and crosstalk that a photonic DNN accelerator would experience, and put in place novel approaches at the device, circuit, and architecture layers to counteract their impact in *CrossLight*. The utilization of TED-based thermal crosstalk management allows us to have MRs placed much closer together, which in turn reduces propagation losses. In addition, *CrossLight* considers a combination of TO and EO tuning which enables the reduction of power and EPB as well. The use of EO tuning in our hybrid tuning approach also provides the advantage of lower latencies, which is apparent in the EPB values.

Table III summarizes the average values of EPB (in pJ/bit) and performance-per-watt (in kiloFPS/Watt) of the photonic accelerators as well as the electronic accelerators considered in this work. It can be observed that the best *CrossLight* configuration (*Cross_opt_TED*) achieves significantly lower EPB and higher performance-per-watt values than all of the accelerators considered. Specifically, against Holylight, which is the best out of the two photonic DNN accelerators considered, *CrossLight* achieves 9.5× lower energy-per-bit and 15.9× higher performance-per-watt. Our work demonstrates the effectiveness of cross-layer design of deep learning accelerators with the emerging silicon photonics technology. With the growing maturity of silicon photonic device fabrication in CMOS-compatible processes, it is expected that the energy costs of device tuning, losses, and laser power overheads will go further down, making an even stronger case for considering optical-domain accelerators for deep learning inference.

**Table III: Average EPB and kiloFPS/Watt values across accelerators**

| Accelerator | Avg. EPB (pJ/bit) | Avg. kiloFPS/watt |
|---|---|---|
| P100 | 971.31 | 24.9 |
| IXP 9282 | 5099.68 | 2.39 |
| AMD-TR | 5831.18 | 2.09 |
| DaDianNao | 58.33 | 0.65 |
| Edge TPU | 697.37 | 17.53 |
| Null Hop | 2727.43 | 4.48 |
| DEAP_CNN | 44453.88 | 0.07 |
| Holylight | 274.13 | 3.3 |
| *Cross_base* | 142.35 | 10.78 |
| *Cross_base_TED* | 92.64 | 16.54 |
| *Cross_opt* | 75.58 | 20.25 |
| *Cross_opt_TED* | 28.78 | 52.59 |

## VI. Conclusion

In this paper, we presented a novel cross-layer optimized photonic neural network accelerator called *CrossLight*. Utilizing silicon photonic device-level fabrication-driven optimizations along with circuit-level and architecture-level optimizations, we demonstrated 9.5× lower energy-per-bit and 15.9× higher performance-per-watt compared to state-of-the-art photonic DNN accelerators. *CrossLight* also shows improvements in these metrics over several CPU, GPU, and custom electronic accelerator platforms considered in our analysis. *CrossLight* shows the promise of cross-layer optimization strategies in countering various challenges such as crosstalk, fabrication-process variations, high laser power, and excessive tuning power. The results presented in this paper demonstrate the promise of photonic DNN accelerators in addressing the need for energy-efficient and high performance-per-watt DNN acceleration.